# Patent Classification by Fine-Tuning BERT Language Model

**Jieh-Sheng Lee** and **Jieh Hsiang**

Department of Computer Science and Information Engineering
National Taiwan University

{d04922013, jhsiang}@ntu.edu.tw

**Abstract**

In this work we focus on fine-tuning a pre-trained BERT model and applying it to patent classification. When applied to large datasets of over two millions patents, our approach outperforms the state of the art by an approach using CNN with word embeddings. In addition, we focus on patent claims without other parts in patent documents. Our contributions include: (1) a new state-of-the-art result based on pre-trained BERT model and fine-tuning for patent classification, (2) a large dataset USPTO-3M at the CPC subclass level with SQL statements that can be used by future researchers, (3) showing that patent claims alone are sufficient for classification task, in contrast to conventional wisdom.

## 1 Introduction

Patent classification is a multi-label classification task. It is challenging because the number of labels can be large, e.g. more than 630 at subclass level. We see this task from two aspects. From the perspective of Deep Learning, pre-training an unsupervised language model on large corpus and fine-tuning the model on downstream tasks have resulted in several state-of-the-art performances recently. Such pre-training models include ELMo (Embeddings from Language Models) [1], ULMFiT (Universal Language Model with Fine-tuning) [2], OpenAI GPT (Generative Pre-Training) [3], BERT (Bidirectional Encoder Representations from Transformers) [4] and OpenAI GPT-2 [5]. Among them, BERT is the most suitable for experiments if having the availability of source code and pre-trained models considered. Therefore, we set a goal to know how well BERT can perform on patent classification after fine-tuning.

From the perspective of patent research, it is time to have a new baseline with a large dataset based on the CPC (Cooperative Patent Classification) system. In general, Deep Learning outperforms other methods when the size of dataset is large. In the past, the sizes of datasets for patent research vary widely. Such variation made comparison difficult. Inference is also less valuable because sometimes the datasets were outdated. In this work, we prepared a new dataset based on the CPC with more than three millions US patents. Patent researchers can leverage the dataset or our approach to cover more tasks, since the entry barriers for data, algorithm and computation are all much lower than before.

The CPC system and the IPC (International Patent Classification) system are two of the most commonly used classification systems. The CPC is a more specific and detailed version of the IPC system. On 1st January 2013, the CPC system came into force and, with which, the United States Patent and Trademark Office (USPTO) replaced its original system. A growing number of national patent offices have decided to follow the CPC [6]. Therefore, it is foreseeable that the CPC system will eventually replace the IPC system as the new standard. However, most of the papers in the field were based on the IPC because of the CLEF-IP competition [7]. The CLEF-IP competition in 2011 was based on the IPC at the subclass level. The dataset consisted of patents filed between 1978 and 2009. Key performances were evaluated with P@1 (precision at top 1), P@5, R@5 (recall at top 5), F1@5 and other metrics. It is not clear to us why R@1 and F1@1 were omitted. This is critical for our work, because a precision value could be very high at the cost of a very low recall. Therefore, P@1 alone might not be a fair number to compare if R@1 and F@1 were not provided. In this work, our metric F1@1 is the best performance of patent classification. We use it to benchmark with the best F1 values in other works, regardless of whether the value is F1@1 or F1@5.

Moreover, our datasets are based on patent claims. The importance of patent claims was underappreciated in the past. When drafting a new patent application, it is a common practice for patent practitioners to draft the patent claims first. The rest of the patent document could be derived or extended from the claims. In patent law, the claims define the scope or the "metes and bounds" of the patented invention. It is a 'bedrock principle' of patent law that 'the claims of a patent define the invention to which the patentee is entitled the right to exclude' [8]. One reason to use patent claims mainly is for our downstream task of patent claim generation in the future. To our knowledge, our work is the first to focus on patent claims and claims only, instead of using claims as supplementary data in the past. To keep our model simpler, we use only the first claim of each patent and leave the benefit of other independent and dependent claims to future research.

## 2  Related Work

We highlight the most relevant works in recent years. Li et al. [9] proposed DeepPatent as a deep learning algorithm based on CNN (Convolutional Neural Network) and word vector embedding. They evaluated the algorithm on the CLEF-IP dataset, compared it with other algorithms in the CLEF-IP competition and claimed a precision of 83.98%, which outperformed all other algorithms. DeepPatent was further tested on USPTO-2M, a newly contributed dataset having 2,000,147 US utility patents in 637 categories at the IPC subclass level after data cleaning. DeepPatent achieved a precision of 73.88% @ Top 1, with no F1@ 1 disclosed. Further experiments by the authors using the same dataset showed that DeepPatent outperformed Random Forest, Decision Tree, BP Networks and Naive Bayes. The best F1 is about 43% @ Top 5. In this work, we use DeepPatent as the baseline to benchmark. We also assumed that the aforementioned methods benchmarked with DeepPatent are unlikely to perform better if the dataset is larger than USPTO-2M.

The idea of fine-tuning a pre-trained language model for patent classification was proposed in the Australasian Language Technology Association Workshop 2018 [10]. The task is to classify Australian patents at the IPC section level (8 labels). The dataset has 75,250 patents (60/40 as training/testing data split). Hepburn [11] used SVM and ULMFiT to achieve the best results in the student category. ULMFiT is a transfer learning technique and the fine-tuning idea is similar to fine-tuning a pre-trained BERT model. The major difference between Hepburn's work and ours is the pre-trained model itself. Another difference in this work is the size of the dataset. Our dataset has over three millions patents while the dataset for the workshop has only 75,250. Our F1 of 80.98% at the CPC section level is also better than the F1 of 78.4% at the IPC section level in Hepburn's work.

It is noted that most of patent classification tasks were done on IPC in the past. Tran and Kavuluru [12] claimed being the first to reinitiate the patent classification task under the new CPC coding scheme. They used logistic regression as a base classifier and exploited extra data or method, such as the hierarchical taxonomy of CPC, the citation records of a test patent, various label ranking and cut-off methods. By experimenting on 436,993 U.S. patents (2010~2011 & 70/30 as training/testing data split) at the subclasses level, their best method achieved 69.89% in micro-F1 score. In this work, we will skip benchmarking with their result since we target a larger dataset without ad-hoc feature engineering. Feature engineering is difficult to scale up in general.

By aiming at CPC and a larger dataset, we also skip any benchmark with the CLEF-IP results. It is uncertain whether fitting a big model like BERT to smaller datasets may make any sense. Conversely it should be more fruitful for other algorithms to benchmark with a larger CPC dataset in the future. Nevertheless, some of the recent works based the legacy CLEF-IP are still noteworthy, for knowing the highest F1 value in the past. For example, comparing with fastText, Yadrintsev et al. [13] claimed that KNN is a viable alternative to traditional text classifiers. Their dataset has 699,000 patents (70/30 as training/testing data split). Their best result achieved 71.02% in micro-F1 score at the IPC subclass level.

Lim and Kwon [14] showed 87.2% precision when using titles, abstracts, claims, technical fields and backgrounds of patents. However, no respective recall or F1 value was disclosed. A fair comparison is therefore not feasible. Their dataset has 564,793 Korean patents at the IPC subclass level. Their method combined multinomial naive Bayes with other tricks such as a Korean language morphological analyzer, using 1,860 stopwords removed and TF-ICF (a variation of the well-known TF-IDF).

Hu et al. [15] showed that a hierarchical feature extraction model can capture both local features as well as global semantics. An n-gram feature extractor based on CNN was designed to extract local features. A bidirectional long–short-term memory (BiLSTM) neural network model was proposed to capture sequential correlations from higher-level representations. The training, validation and test datasets contain 72,532, 18,133, and 2679 mechanical patents from the CLEF-IP dataset. The number of labels is 96 for mechanical patents only. The hierarchical model outperformed other models using CNN, LSTM or BiLSTM alone. Their best F1 is 63.97% @ Top 1. Back to the CLEF-IP competition itself, Verberne and D'hondt [16] reached their best F1-value 70.59% in a series of classification experiments with the Linguistic Classification System (LCS). The training dataset has 905,458 patents and the testing has only 1,000 patents.

## 3  Data

Most of past patent datasets were from the CELF-IP or patent offices. We found it easier to leverage the Google Patents Public Datasets [17] on BigQuery released in 2017. A dataset based on SQL lowers the entry barrier of data preparation significantly. We deem SQL statements as a better way than sharing conventional datasets for two reasons: (1) Separation of concerns. If a dataset contains pre-processing or post-processing already, it could be harder for other researchers to reuse when needing different manipulations. (2) Clarity and flexibility. An SQL statement is precise and easy to revise for different criteria. The SQL statement for our training dataset is listed in the Appendix A.

Our new dataset is called USPTO-3M (3,050,615 patents). Based on the SQL statements, it would be easy for other researchers to cover all patents if computing resource for training is not a constraint. When benchmarking with the DeepPatent, we use its dataset USPTO-2M to benchmark when feasible. If not feasible, we combine other data from USPTO-3M. For example, USPTO-2M does not have claims. In order to see how claims impact performance, we'd combine both datasets. We explain more details in Table 1 of the following section.

## 4  Method & Experimental Setup

In this work, we leverage the released BERT-Base pre-trained model (Uncased: 12-layer, 768-hidden, 12-heads, 110M parameters) [18]. We leave other models such as the BERT-Large (340M parameters) to the future, because the BERT-Base is already sufficient to outperform DeepPatent.

Our implementation follows the fine-tuning example released in the BERT project. For multi-label purpose, we use sigmoid cross entropy with logits function to replace the original softmax function which is suitable for one-hot classification only. We intentionally keep the code change as minimal as possible so as to make the BERT test a vanilla baseline for future experiments to compare against. All hyperparameters remain as default values, e.g. max_seq_length as 128. During our experiments we also observed that it might be sufficient for the max_seq_length to be shorter if having fewer labels, e.g. 9 labels at CPC section level. We leave testing different hyperparameters at different CPC levels to the future.

## 5  Results: PatentBERT vs DeepPatent

In Table 1 we show the original DeepPatent performance in row (a) ~ (d) and our results in row (e) ~ (l). In row (a), DeepPatent achieved the precision of 83.98% @ Top 1 based on EPO and WIPO data. It was claimed that DeepPatent outperforms the state-of-the-art 82.1% @ Top 1 achieved by SVM with full content information of the patent and complicated human-designed features. In row (b), DeepPatent achieved the precision of 73.88% @ Top 1 based on USPTO-2M at the IPC subclass level. It is noted that no recall or F1 were provided in rows (a) and (b). In row (c), DeepPatent achieved the highest F1 of 55.09% @ Top 4 based on EPO and WIPO data.

|   | Method | Patent Data[1] | Train[2] | Test[3] | F1 (%) | Precision (%) | Recall (%) | TREC EVAL |
|---|---|---|---|---|---|---|---|---|
| (a) | DeepPatent | IPC+Title+Abstract | **EPO+WIPO** | EPO | N/A | 83.98 | N/A | Top 1 |
| (b) | DeepPatent | IPC+Title+Abstract | **2006~2014** | 2015-A | N/A | 73.88 | N/A | Top 1 |
| (c) | DeepPatent | IPC+Title+Abstract | EPO+WIPO | EPO | 55.09 | 45.79 | 75.46 | **Top 4** |
| (d) | **DeepPatent** | IPC+Title+Abstract | 2006~2014 | 2015-A | < 45 | < 35 | < 74 | Top 5 |
| (e) | **PatentBERT** | IPC+Title+Abstract | 2006~2014 | 2015-A | 46.85 | 32.19 | 86.06 | Top 5 |
| (f) | PatentBERT | IPC+Title+Abstract | 2006~2014 | 2015-A | 64.91 | 80.61 | 54.33 | **Top 1** |
| (g) | PatentBERT | IPC+**Claim** | 2006~2014 | 2015-A | 63.74 | 79.14 | 53.36 | Top 1 |
| (h) | PatentBERT | **CPC**+Claim | 2006~2014 | 2015-A | 66.83 | 84.26 | 55.38 | Top 1 |
| (i) | PatentBERT | CPC+Claim | 2006~2014 | **2015-B** | 66.80 | 84.24 | 55.35 | Top 1 |
| (j) | PatentBERT | CPC+Claim | **2000~2014** | 2015-B | 66.71 | 84.95 | 54.92 | Top 1 |
| (k) | PatentBERT | CPC+Claim | 2000~2014 | **2016** | 65.89 | 84.89 | 53.84 | Top 1 |
| (l) | PatentBERT | CPC+Claim | 2000~2014 | **2017** | 65.35 | 83.97 | 53.49 | Top 1 |

(1) IPC subclass level: 632 labels. CPC subclass level: 656 labels
(2) Training dataset size:
- EPO: 580,546 patents. WIPO: 161,551 patents.
- USPTO-2M: 2,000,147 patents by the DeepPatent (2006~2015, from USPTO)
- USPTO-3M: 3,050,615 patents, our new dataset with SQL statements (2000~2015, from Google Patents Public Datasets on BigQuery)
- 2006~2014: 1,950,247 patents out of USPTO-2M for DeepPatent. 1,933,105 patents for PatentBERT. Minor discrepancy exists due to different data sources and probably preprocessing criteria.
- 2000~2014 : 2,900,615 patents out of USPTO-3M for PatentBERT

(3) Testing dataset:
- EPO: 1,350 patents
- 2015-A: 49,900 patents out of USPTO-2M for DeepPatent. 49,670 patents for PatentBERT (out of USPTO-3M and based on DeepPatent's list of test patents)
- 2015-B: 150,000 of the 298,559 patents in 2015 (from USPTO-3M)
- 2016: 150,000 of the 298,559 patents in 2016 (from BigQuery)
- 2017: 150,000 of the 298,559 patents in 2017-01~2017-08 (from BigQuery)

Table 1: Patent Classification Performance

In row (d), we list the ranges of F1, precision and recall because no precise numbers were provided in their work. Based on USPTO-2M dataset, the highest F1 value achieved by DeepPatent is lower than 45% @Top 5.

PatentBERT results start from row (e). We compare row (e) with row (d) to show that PatentBERT reaches a higher F1 @ Top 5 as 46.85%. In row (f), the F1 value reaches even higher as 64.91% @ Top 1. In row (g), we use patent claims to replace title and abstract. The F1 value drops a little but the difference does not matter to our future downstream task. In row (h), we show that CPC is better than IPC at subclass level and the F1 value reaches 66.83%. It is noted that the precision value 84.26% achieved by PatentBERT outperforms DeepPatent in both row (a) and (b).

In row (i), we show that the F1 value remains stable when the size of test dataset triples. In row (j), we show that the F1 value is also stable after using the larger dataset USPTO-3M. We further observed from row (j), (k) and (l) that the F1, precision and recall values all dropped slightly as the date (year) of the test data moved further away from the training data.

# 6 Conclusion

Patents might be an ideal data source for human to solve *artificial innovation*. However, patent classification as groundwork has been a challenging task with no satisfactory performance for decades. In this paper we present a new state-of-the-art approach based on fine-tuning a pre-trained BERT model and it outperforms DeepPatent. Our results also show that using patent claims alone is sufficient for classification task. Most important of all, the recent success of the two-stage framework (pre-training & fine-tuning) in Deep Learning is promising for patent researchers to explore more in the future. Patent classification in this work is just an example.

## Appendix A

- The following SQL selects the first claims of all US utility patents in 2013 and aggregates the CPC codes at subclass level: (data source: Google Patents Public Datasets on BigQuery)

SELECT STRING_AGG(distinct t2. group_id order by t2. group_id) AS cpc_ids, t1.id, t1.date, text

FROM `patents-public-data.patentsview.patent` t1,

`patents-public-data.patentsview.cpc_current` t2,

`patents-public-data.patentsview.claim` t3

where t1.id = t2.patent_id

and t1.id = t3.patent_id

and timestamp(t1.date) >= timestamp('2013-01-01')

and timestamp(t1.date) <= timestamp('2013-12-31')

and t3.sequence='1'

and t1.type='utility'

group by t1.id, t1.date, t3.text